**Natural language syntax complies with the free-energy principle**

Elliot Murphy[1,2], Emma Holmes[3,4], Karl Friston[4]

1. Vivian L. Smith Department of Neurosurgery, McGovern Medical School, University of Texas Health Science Center, Houston, Texas, 77030, USA

2. Texas Institute for Restorative Neurotechnologies, University of Texas Health Science Center, Houston, Texas, 77030, USA

3. Department of Speech Hearing and Phonetic Sciences, University College London, London, WC1N 1PF, UK

4. The Wellcome Centre for Human Neuroimaging, UCL Queen Square Institute of Neurology, London, WC1N 3AR, UK

Correspondence concerning this article should be addressed to Elliot Murphy (elliot.murphy@uth.tmc.edu).

Word count: 13,443

Figure count: 1

**Abstract:**

Natural language syntax yields an unbounded array of hierarchically structured expressions. We claim that these are used in the service of active inference in accord with the free-energy principle (FEP). While conceptual advances alongside modelling and simulation work have attempted to connect speech segmentation and linguistic communication with the FEP, we extend this program to the underlying computations responsible for generating syntactic objects. We argue that recently proposed principles of economy in language design—such as "minimal search" criteria from theoretical syntax—adhere to the FEP. This affords a greater degree of explanatory power to the FEP—with respect to higher language functions—and offers linguistics a grounding in first principles with respect to computability. We show how both tree-geometric depth and a Kolmogorov complexity estimate (recruiting a Lempel–Ziv compression algorithm) can be used to accurately predict legal operations on syntactic workspaces, directly in line with formulations of variational free energy minimization. This is used to motivate a general principle of language design that we term Turing–Chomsky Compression (TCC). We use TCC to align concerns of linguists with the normative account of self-organization furnished by the FEP, by marshalling evidence from theoretical linguistics and psycholinguistics to ground core principles of efficient syntactic computation within active inference.

Implementational models of language must be plausible from the perspective of neuroanatomy (Embick & Poeppel 2015), but they must also be plausible from the perspective of how biophysical systems behave. We will argue that the structuring influence of the free-energy principle (FEP) can be detected in language, not only via narrative (Bouizegarene et al. 2020), interpersonal dialogue (Friston et al. 2020), cooperative/intentional communication (Vasil et al. 2020) and speech segmentation (Friston et al. 2021), but also at the more fundamental level of what linguists consider to be basic phrase-level computations (Berwick & Stabler 2019, Chomsky 1949, 1951, 1956, 1959, 2021a, 2021b, 2021c).

Natural language syntax yields an unbounded array of hierarchically structured expressions. We argue that many historical insights into syntax are consistent with the FEP—providing a novel perspective under which the principles governing syntax are not limited to language, but rather reflect domain-general processes. This is consistent with a strain within theoretical linguistics that explores how syntactic computation may adhere to "general principles that may well fall within extra-biological natural law, particularly considerations of minimal computation" (Chomsky 2011: 263), such that certain linguistic theories might be engaging with general properties of organic systems (Chomsky 2004, 2014). Here, we consider the idea that many aspects of natural language syntax may be special cases of a variational principle of least free-energy. To this end, we examine whether a complexity measure relevant to formulations of free-energy—namely, Kolmogorov complexity (Hutter 2006, MacKay 1995, Wallace & Dowe 1999)—relates to legal syntactic processes.

While the FEP has a substantial explanatory scope, across a large range of cognitive systems, it can also be seen as a method or principle of least action for multi-disciplinary research (Ramstead et al. 2021), in much the same way that the notion of economy is typically entertained in linguistics as a *programmatic* notion (Chomsky 1995). The FEP itself has been argued to be more of a conceptual-mathematical model for self-organizing systems (for some, it is a "generic" model; Barandiaran & Chemero 2009), or a guiding framework (Colombo & Wright 2021). Thus, when we argue that natural language syntax "complies" with the FEP, this is not to imply that the FEP necessarily bears any specific, direct predictions for linguistic behaviour. Rather, it motivates the construction of conceptual arguments for how some property of organic systems might be seen as realizing the FEP.

We begin by summarising the FEP, and describe how syntactic principles are consistent with it. We consider how the FEP is a variational principle of "least action", such as those that describe systems with conserved quantities (Coopersmith 2017). We then review key observations from linguistics that speak to the structuring influence of computational efficiency, involving "least effort" and "minimal search" restrictions (Bošković & Lasnik 2007, Gallego & Martin 2018, Larson 2015), viewing language as a product of an individual's mind/brain, following the standard 'I-language' (Chomsky 1986, 2000) perspective in generative linguistics (i.e., 'internal', 'individual', 'intensional'). After modeling the complexity of postulated minimal search procedures—versus their ungrammatical alternatives across a small but representative number of exemplar cases—we propose a unifying principle for how the goals of the FEP might be realised during the derivation of syntactic structures, which we term Turing–Chomsky Compression. We conclude by highlighting directions for future work.

## 1. Active Inference and the Free-Energy Principle

Before we evaluate any work pertaining to linguistic behaviour, this section introduces key elements of the FEP that motivate its application to language.

### 1.1. The Free-Energy Principle

The FEP has a long history in theoretical neuroscience (see Friston 2010 for a review). It states that any adaptive change in the brain will minimise free-energy, either over evolutionary time or immediate, perceptual time (Ramstead et al. 2018). Free-energy is an information-theoretic quantity and is a function of sensory data and brain states: in brief, it is the upper bound on the 'surprise'—or surprisal (Tribus 1961)—of sensory data, given predictions that are based on an internal model of how those data were generated. The difference between free-energy and surprise is the difference (specified by the Kullback-Leibler divergence) between probabilistic representations encoded by the brain and the true conditional distribution of the causes of sensory input. This is evident in the following equation, which specifies variational free energy ($F$) as the negative log probability of observations ($\tilde{o}$) under a generative model (i.e., 'surprise') plus the Kullback–Leibler divergence ($D_{KL}$) between the approximate posterior distribution and the true posterior distribution (where $Q$ indicates posterior beliefs, $\tilde{s}$ indicates the states in the generative model, and $P$ indicates the probability under the internal model):

$$F = -\ln P(\tilde{o}) + D_{KL}[Q(\tilde{s}) \,\|\, P(\tilde{s} \mid \tilde{o})]$$

(Eq. 1)

Unlike surprise itself, variational free energy can be evaluated (for a detailed explanation, see Friston, FitzGerald et al. 2017). Under simplifying assumptions, free-energy can be considered as the amount of prediction error; for a mathematical comparison, see Friston, Parr et al. (2017). Minimising free energy minimises surprise, and is equivalent to maximising the evidence for the internal model of how sensory data were generated. By minimising free-energy, the brain is essentially performing approximate Bayesian inference. By reformulating variational free energy—in a way that is mathematically equivalent; see Penny et al. (2004), Winn & Bishop (2005)—we see that free-energy can be considered as a trade-off between accuracy and complexity, whereby the best internal model is the one that accurately describes the data in the simplest manner (where $E_Q$ indicates the expected value, and the other variables are the same as those defined above):

$$
\begin{aligned}
F &= E_Q[\ln Q(\tilde{s}) - \ln P(\tilde{o} \mid \tilde{s}) - \ln P(\tilde{s})] \\
&= E_Q[\ln Q(\tilde{s}) - \ln P(\tilde{s} \mid \tilde{o}) - \ln P(\tilde{o})] \\
&= \underbrace{D_{KL}[Q(\tilde{s}) \parallel P(\tilde{s} \mid \tilde{o})]}_{\text{relative entropy}} - \underbrace{\ln P(\tilde{o})}_{\text{log evidence}} \\
&= \underbrace{D_{KL}[Q(\tilde{s}) \parallel P(\tilde{s})]}_{\text{complexity}} - \underbrace{E_Q[\ln P(\tilde{o} \mid \tilde{s})]}_{\text{accuracy}}
\end{aligned}
$$

(Eq. 2)

Because the relative entropy can never be less than zero, the variational free energy provides an upper bound on negative log evidence: equivalently, the negative free energy provides a lower bound on log evidence; known as an evidence lower bound (ELBO) in machine learning (Winn & Bishop 2005). The final equality shows a complementary decomposition of variational free energy into accuracy and complexity. In effect, it reflects the degree of belief updating afforded by some new sensory data; in other words, how much some new evidence causes one to "change one's mind". A good generative model—with the right kind of priors—will minimise the need for extensive belief updating and thereby minimise complexity.

The complexity part of variational free energy will become important later, when we will be evaluating the complexity of syntactic processes using a measure derived both in spirit and mathematical heritage from the foundations of the FEP. To present some initial details about this, consider how complexity also appears in treatments of universal computation (Hutter 2006) and in the guise of minimum message or description lengths (Wallace & Dowe 1999). Indeed, in machine learning, variational free energy minimisation has been cast as minimising complexity—or maximising efficiency in this setting (MacKay 1995). One sees that same theme emerge in predictive coding—and related—formulations of free energy minimisation, where the underlying theme is to compress representations (Schmidhuber 2010), thereby maximising their efficiency (Barlow 1961, Linsker 1990, Rao & Ballard 1999). We will return to these topics below when we begin to formalise features of natural language syntax.

Lastly, the FEP can also be considered from the perspective of a Markov blanket (see Kirchhoff et al. 2018, Palacios et al. 2020, Parr et al. 2020 for detailed explanation), which instantiates a statistical boundary between internal states and external states. In other words, internal (e.g., in the brain) and external (e.g., in the external world) states are conditionally independent: they can only influence one another through blanket states. The blanket states can be partitioned into sensory states and active states. External states affect internal states only through sensory states, while internal states affect external states only through active states. The implicit circular causality is formally identical to the perception-action cycle (Fuster 2004). Under previous accounts (Friston, FitzGerald et al. 2017), the brain can minimise free-energy either through perception or action. The former refers to optimising (i.e., using approximate Bayesian inference to invert) its probabilistic generative model that specifies how hidden states cause sensory data; in other words, inferring the cause of sensory consequences by minimising variational free energy. The latter refers to initiating actions to sample data that are predicted by its model—which we turn to next.

### 1.2. Active Inference

The enactive component of active inference rests on the assumption that action is biased to realize preferred outcomes. Beliefs about which actions are best to pursue rely on predictions about future outcomes, and the probability of pursuing any particular outcome is given by the *expected free energy* of that action. Expected free energy ($G$) can be expressed as the combination of extrinsic and epistemic value (Friston, Parr et al. 2017), where $\pi$ is a series of actions (i.e., the policy) being pursued, $\tau$ is the current time point, and the other variables are the same as those defined above:

$$G(\pi) = \sum_{\tau} G(\pi, \tau)$$

$$G(\pi, \tau) = E_Q[\ln Q(s_\tau \mid \pi) - \ln Q(s_\tau \mid o_\tau, \pi) - \ln P(o_\tau)]$$

$$= \underbrace{E_Q[\ln Q(s_\tau \mid \pi) - \ln Q(s_\tau \mid o_\tau, \pi)]}_{\text{(negative) mutual information}} - \underbrace{E_Q[\ln P(o_\tau)]}_{\text{expected log evidence}}$$

$$= \underbrace{E_Q[\ln Q(o_\tau \mid \pi) - \ln Q(o_\tau \mid s_\tau, \pi)]}_{\text{(negative) epistemic value}} - \underbrace{E_Q[\ln P(o_\tau)]}_{\text{extrinsic value}}$$

(Eq. 3)

Extrinsic value is the expected evidence for the internal model under a particular course of action, whereas epistemic value is the expected information gain: in other words, the extent a series of actions reduces uncertainty.

Notice that the expected versions of relative entropy and log evidence in the free energy (Eq. 2) now become intrinsic and extrinsic value respectively (Eq. 3). As such, selecting an action to minimise expected free energy reduces expected surprise (i.e., uncertainty) in virtue of maximising the information gain while—at the same time—maximising the expected log evidence; namely, actively self-evidencing (Hohwy 2016, 2020). When applied to a variety of topics in cognitive neuroscience, active inference has been shown to predict human behaviour and neuronal responses; e.g., Brown et al. (2013), Friston, Lin et al. (2017), Friston, Rosch et al. (2017), Parr and Friston (2017, 2019), Smith et al. (2019).

As will soon become clear, we will be using these observations concerning complexity to motivate a specific application of these ideas to natural language syntax, utilizing a measurement of algorithmic complexity that shares a mathematical heritage with the FEP, as outlined here.

### 1.3. Belief Updating

Belief updating refers to a process by which free energy is minimised. By specifying a process theory that explains neuronal responses during perception and action, neuronal dynamics have previously been cast as a gradient flow on free energy (known as variational free energy in physics, introduced in Feynman 1972; see Hinton & Zemel 1994); we refer the reader to Friston, FitzGerald et al. (2017) for a treatment of neuronal message passing and belief propagation. That is to say, any neural process can be formulated as a minimisation of the same quantity used in approximate Bayesian inference (Andrews 2021, Hohwy 2016). We provide an example of the computational architecture implied by this formulation of belief updating in the brain (Fig. 1). This illustrative example is based upon a discrete state space generative model, where the equations describe the solutions to Bayesian belief updating of expectations pertaining to hidden states, policies, policy precision and parameters.

In short, the brain seeks to minimise free energy, which is mathematically equivalent to maximising model evidence. This view of neuronal responses can be conceived with respect to Hamilton's principle of least action, whereby action is the path integral of free energy.

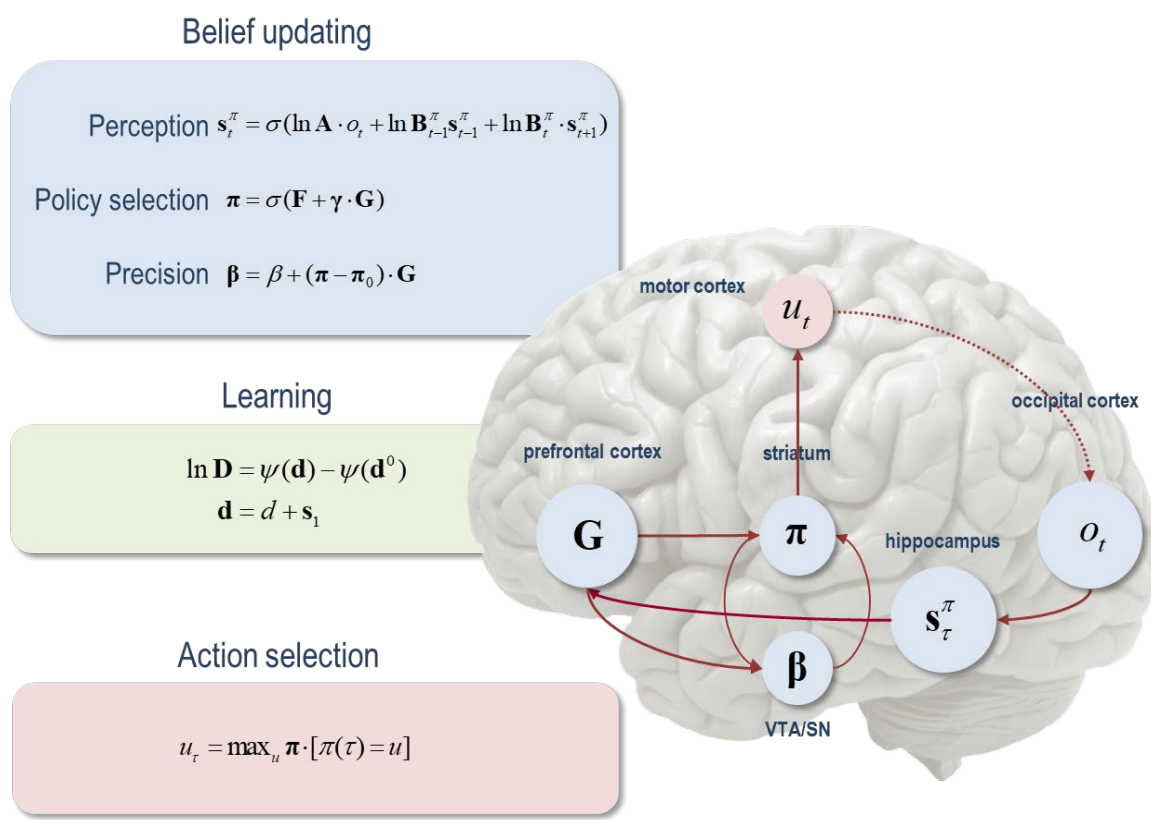

## Functional anatomy and message passing

**Belief updating**

Perception $\mathbf{s}_t^\pi = \sigma(\ln \mathbf{A} \cdot o_t + \ln \mathbf{B}_{t-1}^\pi \mathbf{s}_{t-1}^\pi + \ln \mathbf{B}_t^\pi \cdot \mathbf{s}_{t+1}^\pi)$

Policy selection $\boldsymbol{\pi} = \sigma(\mathbf{F} + \boldsymbol{\gamma} \cdot \mathbf{G})$

Precision $\boldsymbol{\beta} = \beta + (\boldsymbol{\pi} - \boldsymbol{\pi}_0) \cdot \mathbf{G}$

**Learning**

$\ln \mathbf{D} = \psi(\mathbf{d}) - \psi(\mathbf{d}^0)$
$\mathbf{d} = d + \mathbf{s}_1$

**Action selection**

$u_\tau = \max_u \boldsymbol{\pi} \cdot [\pi(\tau) = u]$

motor cortex $u_t$

occipital cortex

prefrontal cortex    striatum

$\mathbf{G}$    $\boldsymbol{\pi}$    hippocampus    $o_t$

$\boldsymbol{\beta}$    $\mathbf{s}_\tau^\pi$

VTA/SN

**Figure 1: Schematic overview of belief updates for active inference under discrete Markovian models**. The left panel lists the belief updating equations, associating various updates with action, perception, policy selection, precision and learning. The left panel assigns the variables (sufficient statistics or expectations) that are updated to various brain areas. This attribution serves to illustrate a rough functional anatomy—implied by the form of the belief updates. In this simplified scheme, we have assigned observed outcomes to visual representations in the occipital cortex and updates to hidden states to the hippocampal formation. The evaluation of policies, in terms of their (expected) free energy, has been placed in the ventral prefrontal cortex. Expectations about policies *per se* and the precision of these beliefs have been attributed to striatal and ventral tegmental areas to indicate a putative role for dopamine in encoding precision. Finally, beliefs about policies are used to create Bayesian model averages over future states, which are fulfilled by action. The red arrows denote message passing. In brief, the parameters of the generative model correspond to matrices or arrays encoding the likelihood **A**, prior state transitions **B**, and initial hidden states **D**. **F** corresponds to the free energy of each policy and **G** corresponds to the expected free energy, which is weighted by a precision or softmax parameter γ that is usually attributed to dopaminergic neurotransmission. See Friston, FitzGerald et al. (2017) for further explanation of the equations and variables.

## 1.4.  Previous Applications

Applying active inference to language relies on finding the right sort of generative model, and many different structures and forms of generative model are possible. Most relevant to the current application, deep temporal models accommodate a nesting of states that unfold at different temporal scales. Since language output is inherently temporal, this leads to the question of how to map hierarchical structures onto serial outputs (Epstein et al. 1998), and models that are deep in time allow us to deconstruct associated nested structures.

Recently, a deep temporal model for communication was developed based on a simulated conversation between two synthetic subjects, showing that certain behavioural and neurophysiological correlates of communication arise under variational message passing (Friston et al. 2020). The model incorporates various levels that operate at different temporal scales. At the lowest level, it specifies mappings among syntactic units that, when combined with particular semantic beliefs, predict individual words. At a higher level (i.e., longer temporal scales), the model contains beliefs about the context it is in, which specifies the syntactic structure at the level below. This model is congruent with core assumptions from linguistics concerning the generative nature of language. Specifically, elementary syntactic units provide *belief structures* that are used to reduce uncertainty about the world, through rapid and reflexive categorization of events, objects and their relations. Then, sentential representations can be thought of as structures designed (partially) to consolidate and appropriately frame experiences, and to contextualise and anticipate future experiences. The range of parseable syntactic structures available to comprehenders provides alternate hypotheses that afford parsimonious explanations for sensory data and, as such, preclude overfitting (Sengupta & Friston 2018). If the complexities of linguistic stimuli can be efficiently mapped to a small series of regular syntactic formats, this contributes to the brain's goal of restricting itself to a limited number of states. Intuitively, knowing that you can only have said this or that greatly simplifies the inference problem, rendering exchanges between like-minded artefacts more efficient and economic.

Before moving forward, we stress here that we will be working within the framework of theoretical linguistics (which deals with *derivational* stages of word-by-word, element-wise operations that underlie sentences), and not a framework such as corpus linguistics (which deals with the *output* of the generative/derivational process).

Questions we raise here therefore cannot be addressed by consulting large corpora, but instead require investigation of incremental computational steps that ultimately appear to be responsible for the complex forms of human language behavior studied by sociolinguistics, corpus linguistics, and historical linguistics. Relatedly, embracing the traditional distinction between competence and performance, our focus will be on the former (the mental formatting and generation of linguistic structure), and not on the range of complex cognitive processes that enter into the use of language in a specific context. The FEP has many implications for the study of language, and we focus here on a very narrow, specific topic of derivational syntactic complexity (language knowledge, as opposed to language use). Whether or not free energy is minimized across corpora, or across paragraphs from clippings of newspaper columns, or across the world's languages, is an open topic—but here we will focus on the computational system of human language.

## 2. Computational Principles and Syntactic Hierarchies

### 2.1.   A System of Discrete Infinity

How can the FEP contribute to our understanding of syntactic computation? Most immediately, it provides fundamental constraints on the physical realisation of a computational system. Consider first the three principles in classical recursive function theory which allow functions to compose (Kleene 1952): substitution; primitive recursion; and minimisation. These are all designed in a way that one might think of as computationally efficient: they reuse the output of earlier computations. Substitution replaces the argument of a function with another function (possibly the original one);

primitive recursion defines a new function on the basis of a recursive call to itself, bottoming out in a previously defined function (Lobina 2017); minimisation (also termed 'bounded search') produces the output of a function with the smallest number of steps (see also Piantadosi 2021, for whom human thought is essentially like Church encoding). More broadly, we note that free energy minimization—by construction—entails Bayesian inference, which in turn is a computational process, and so the FEP entails computationalism (Korbak 2021) and at least a type of (basic) computational architecture for language we assume here (but see Kirchhoff & Robertson 2018). Examining some core principles of recursion, natural language clearly exhibits minimisation, while binary branching of structures (Radford 2016) limits redundant computation, reducing the range of possible computations. Even limitations on short-term memory have been hypothesized to contribute to the efficiency of memory search (MacGregor 1987).

Syntax involves the construction of binary-branching hierarchically organized sets via the operation MERGE, taking objects from the lexicon *or* objects already part of the workspace. For example, given the set {X, Y}, we can either select a new lexical object and MERGE it, to form {Z, {X, Y}}, or we can select an existing object and MERGE it to the same workspace, to form {X, {X, Y}}[1]. MERGE serves a similar role to an elementary function, as in the theory of computability (e.g., the zero function, the identity function), in that it is meant to be non-decomposable. Putting many subsidiary technical details aside, these sets are then 'labeled' and given a syntactic identity, or a 'head' (Frampton & Gutmann 1999, Murphy 2015a, Murphy, Woolnough et al. 2022), based on which element is most structurally prominent and easiest to search for (i.e., Z in the structure $\{Z, \{X, Y\}\}$)[2]. For example, 'the old man' is a Determiner Phrase (DP), and has the distributional properties of a DP, since 'the' is equivalent to Z, and 'old man' equivalent to $\{X, Y\}$. Labeling takes place when conceptual systems access the structures generated by syntax. This occurs at distinct derivational punctuations based on the configuration and subcategorization demands of the lexical items involved (e.g., in many instances subjects seem to be featurally richer than objects, and provide the relevant feature for the label; Longobardi 2008). For example, in the case of head-complement structures this is done immediately after MERGE (Bošković 2016). MERGE can also derive some set-theoretic properties of linguistic relations, such as *membership*, *dominate* and *term-of*, as well as the derived relation of *c-command* (=sister of) which is relevant for interpreting hierarchical relations between linguistic elements (Haegeman 1994)[3]. These also appear to be the simplest possible formal relations entertained, potentially indexing a feature of organic computation that adheres closely to criteria of simplicity (Chomsky 2022).

---

[1] There are recent debates concerning whether merging new lexical objects is more computationally demanding (since it requires searching the lexicon) than merging lexical objects that are already in the workspace. This may motivate a Move-over-Merge bias (reversing the Merge-over-Move assumption from the early minimalism of the 1990s), however we leave this issue to one side.

---

[2] There is increasing evidence that only elements in the workspace that have been labeled can be subject to movement (Bošković 2021).

[3] Note that there are also numerous cases in which principles previously seen as purely syntactic have been re-framed as, for example, phonological in nature (e.g., string adjacency; Bobaljik 2002; see also Richards 2016, Samuels 2011).

One might also think of MERGE as physical information coarse-graining (i.e., the removal of superfluous degrees of freedom in order to describe a given physical system at a different scale), with the core process of syntax being information renormalization according to different timescales. For instance, MERGE can be framed as a probability tensor implementing coarse-graining, akin to a probabilistic context-free grammar (Gallego & Orús 2022). The model proposed in Gallego and Orús (2022: 20) assumes that language is "the cheapest non-trivial computational system", exhibiting a high degree of efficiency with respect to its MERGE-based coarse-graining.

Natural language syntax exhibits *discrete units* which lead to a *discreteness-continuity duality* (the boundary between syntactic categories can be non-distinct)[4]. Syntax is driven by *closeness of computation* (syntactic objects X and Y form a distinct syntactic object, {X, Y}). Its objects are *bounded* (a fixed list, e.g., N, V, Adj, Adv, P, C, T, *n*, *v*, Asp, Cl, Neg, Q, Det) and their hierarchical ordering is based on a specific functional sequence such as C-T-*v*-V (e.g., C is always higher than V; Starke 2004) which imposes direct restrictions on combinatorics (Adger & Svenonius 2011). These objects can be combined in *cycles* (Frampton & Gutmann 1999), which can be extended to form *non-local dependencies*. As we will discuss, these properties are guided by principles of minimal search (an optimal tree-search procedure, informed by notions from computer science; Aycock 2020, Ke 2019, Roberts 2019) and least effort (Larson 2015), fulfilling the imperatives of active inference to construct meaningful representations as efficiently as possible (Bouizegarene et al. 2020), contributing to surprise minimisation.

---

[4] In active inference, the use of discrete—as opposed to continuous—states in generative models is an enormously potent way of minimising complexity. For example, if it is sufficient to carve the world (i.e., the causes of my sensations) into a small number of hidden states, one can minimise the complexity of belief updating by not redundantly representing all the fine-grained structure within any one state. Similarly, factorisation plays a key role in structuring our hypotheses or expectations that provide the best explanation for sensations. Perhaps the clearest example here is the carving of a sensory world into *what* and *where*. This means one does not have to represent the location of every possible object in one's visual cortical hierarchy, just *where* an object is and *what* an object is—and then integrate these representations to explain visual input (Ungerleider & Haxby 1994).

## 2.2.  Compositionality

Recently, certain efficiency principles at the conceptual interface (where syntax interfaces with general conceptualization) have been proposed (Pietroski 2018), such that the 'instructions' that language provides for the brain to build specific meanings are interpreted with notable simplicity. Leaving more technical details aside, this is ultimately achieved (in Pietroski's model) through M-join (e.g., $F(\_) + G(\_) \rightarrow F^{\wedge}G(\_)$, which combines *monadic* concepts, like *red + boat*) and D-join (e.g., $D(\_,\_) + M(\_) \rightarrow \exists[D(\_,\_)^{\wedge}M(\_)]$, which merges *dyadic* concepts with monadic concepts, deriving the meaning of *X verb(ed) Y*). Hence, natural language permits limited dyadicity as a very minimal departure from the most elementary monadic combinatorial system. Human language is marginally more expressive (in its conceptual interpretations) than a first-order language (i.e., one set, and one embedding), but the interpretation system is the least complex needed to express dyadicity and permit *relations* between sets. As with models of syntax invoking a simple process of binary set-formation to derive recursion, by restricting the number of computational procedures able to generate semantic structures, this model (and others like it; Heim & Kratzer 1998) restricts in highly predictable ways the possible range of outputs. This model also assumes that there are 'fetchable' concepts that language can use for compositionality, and non-fetchable concepts. For instance, language seems to make considerable use of certain contentful concepts (e.g., evidentiality) but not others (e.g., colour, worry/sollicitativity) (Adger 2019, Peterson 2016).

Consider also how, in neo-Davidsonian event semantics, conjunction is limited to predicates of certain semantic types (Schein 1993, Pietroski 2005). Certain semantic rules of composition, in (1b), have been claimed to arise directly from more elementary syntactic computations (Pietroski 2018) which adhere to principles of efficient computation.

(1) a.     Dutch shot Micah quickly.

    b.     ∃e[Agent(e, Dutch) & Theme(e, Micah) & shot(e) & quickly(e)]

In this connection, it has further been observed that language acts as an artificial context which helps "constrain what representations are recruited and what impact they have on reasoning and inference" (Lupyan & Clark 2015: 283). Words themselves are "highly flexible (and metabolically cheap) sources of priors throughout the neural hierarchy" (Ibid) (for discussion of simplicity in semantic computation, see Al-Mutairi 2014, Bošković & Messick 2017, Collins 2020, Gallego & Martin 2018, González Escribano 2005, Hauser et al. 2002, Hornstein & Pietroski 2009). The notion of *truth* is also deeply wedded to complex syntax (Hinzen 2016, Hinzen & Sheehan 2013), and is a notion that enormously increases an agent's ability to generate environmental inferences.

To briefly elaborate on the language/meaning relation, consider how universal syntactic hierarchies such as Complementizer Phrases (CPs) containing Tense

Phrases (TPs), and TPs containing Verb Phrases (VPs) (i.e., [CP[TP[VP]]]), can be thought to emerge directly from extra-linguistic conceptual relations, or the "cognitive proclivity to perceive experience in terms of events, situations, and propositions" (Ramchand & Svenonius 2014: 33). An *event* (VP) is a kind of timeless description (e.g., 'run'), and when an event is merged with a deictic anchor the result is a *situation* (TP), a time-anchored eventuality. Only when a situation is merged with a specific discourse link does it become a proposition (CP), providing speaker perspectives (e.g., 'John will run').

## 3. A Kolmogorov Complexity Estimate for Narrow Syntax

### 3.1. Economy

The notion of simplicity has been a methodological value which has guided linguistic inquiry for decades (Terzian & Corbalan 2021). Chomsky (2021b: 13) notes that measures of simplicity in linguistics have traditionally focused on symbol counting, which transpired to be too crude a measure, and that "measuring simplicity is an essential task and is no simple matter". We aim in this section to elaborate a measure of syntactic complexity that connects directly to the principles that underwrite active inference. We will not be concerned with typological, phonological or acquisitional notions of complexity, which form the bulk of current complexity literature. Instead, we are interested in underlying representational issues that pertain to syntax-semantics. Even the most recent explorations of simplicity in language, such as the volume on simplicity in grammar learning in Katzir et al. (2021), focuses on modelling minimum description length in phonology and morphology, but not processes pertaining to the derivation of syntactic objects.

A number of economy principles have been proposed in theoretical linguistics: the No Tampering Condition (Chomsky 2008), Minimal Link Condition (Chomsky 1995), Minimal Yield (Chomsky 2021c), Extension Condition (Chomsky 1995), Last Resort (Chomsky 1995), Relativised Minimality (Rizzi 1990, 2001), Inclusiveness Condition (Chomsky 1995), Precedence Resolution Principle (Epstein et al. 1998), Scope Economy (Fox 2000), Phase Impenetrability Condition (Chomsky 2004), Full Interpretation (Freidin & Lasnik 2011, Lohndal & Uriagereka 2016), Global Economy Condition (Sternefeld 1997), Feature Economy (van Gelderen 2011), Accord Maximization Principle (Schütze 1997), Input Generalisation (Holmberg & Roberts 2014), Maximise Minimal Means (Biberauer 2019a), Resource Restriction (Chomsky et al. 2019), and Equal Embedding (Murphy & Shim 2020) (for further discussion, see Frampton & Gutmann 1999, Fukui 1996, Titov 2020).

Although economy principles have figured in models of phonology, morphology and the lexicon (e.g., the Elsewhere condition, underspecification) for decades, it is only relatively recently that theories of syntax have embraced economy not just as a heuristic guiding research, but more concretely as a constitutive principle of language design (Leivada & Murphy 2021, Reuland 2011, Sundaresan 2020). These have been framed within a linguistic context, often invoking domain-specific notions (Wilder et al. 1997), despite a core part of the intended project of modern theoretical linguistics being to embed linguistic theory within principles general to cognition. Motivating language-specific computational generalizations by direct reference to the FEP may broaden the explanatory scope for the existence and prevalence of particular syntactic phenomena. Since linguists lack a general theory of computational efficiency for language (e.g., Gallego & Chomsky 2020: "To be sure, we do not have a general theory of computational efficiency"), additional support with respect to grounding these concerns within a well-motivated framework for general organic behaviour will likely prove productive. There are many promising paths to take here: minimising energy expenditure during language processing (Rahman & Kaykobad 2005), shortening description length (Schmidhuber 2015), reducing Kolmogorov complexity (Ming & Vitányi 2008, Wallace & Dowe 1999), and the degree of requisite belief updating; relatedly, we might consult the principles of minimum redundancy and maximum efficiency in perception (Barlow 1961, 1974, Wipf & Rao 2007). We will provide a concrete exploration of one of these notions (Kolmogorov complexity) below.

A core fact about many natural language expressions is that they involve arrangements of nested constituents that enter into relations and dependencies of various kinds. How are syntactic operations compressed into determinate, unambiguous instructions to conceptual systems, and are there any general laws of organic design that can be inferred from the FEP that appear to constrain this process (and which successfully predict which objects *cannot* be constructed)?

Consider how under the No Tampering Condition the merging of two syntactic objects, X and Y, leaves X and Y unchanged. The set {X, Y} created by MERGE (Chomsky et al. 2019) cannot be broken and no new features can be added[5]. The original structure in (2a) can be modified by the merging of a new element, λ, to form (2b), adhering to the No Tampering Condition, while (2c) violates this condition since the original structure (2a) is modified (Lohndal & Uriagereka 2016); hence why adjuncts that merge 'downstairs' do not alter the structure of the object they adjoin to (adjuncts are not labeled; Bošković 2015), i.e., if 'John saw Mary [in the park]', then the fact that John saw Mary does not change (subscripts denote syntactic heads/labels, standard script denotes lexical items, where (2a) could represent a structure like 'the red boat')[6].

(2) a.      [$_\alpha$ [$_\beta$ [$_\gamma$ [$_\delta$ ε]]]]

    b.      [$_\alpha$ λ [$_\alpha$ [$_\beta$ [$_\gamma$ [$_\delta$ ε]]]]]

    c.      [$_\alpha$ [$_\beta$ [$_\gamma$ [$_\delta$ [$_\varepsilon$ ε λ]]]]]

Further, it is more *economical* to expand a structure, as in (2b), than to backtrack and modify a structure that has already been built, as in (2c) (Lasnik & Lohndal 2013). How

---

[5] MERGE has been defined as an operation on a workspace and its objects, formalized as follows (WS = workspace; P/Q = workspace objects such as linguistic features; X = additional elements):

MERGE(P,Q,WS) = WS' = {{P,Q},$X_1$,…,$X_n$}.

Other linguistic frameworks assume some similar, basic structure-building operation, e.g., *Forward-Backward Application* in Combinatory Categorial Grammar (Steedman 2000), *Substitution* in Tree-Adjoining Grammar (Joshi & Schabes 1997). We put aside here some controversies about the relation between a MERGE-based syntax and set theory (see Gärtner 2021).

[6] Adjacent to minimalist syntax, Optimality Theory assumes that gratuitous adjunction to a maximal projection violates an economy condition (SPECLEFT) (Broekhuis & Vogel 2009, Grimshaw 2001). For syntactic economy conditions in Lexical Functional Grammar, see Dalrymple et al. (2015).

can we more formally demonstrate these claims? We turn here to Kolmogorov complexity.

## 3.2. Compression

Kolmogorov complexity is a measure of the length of the shortest program that can reproduce a given pattern (Kolmogorov 1965, Li & Vitányi 2019). While measures of minimum description length and Kolmogorov complexity have been typically applied to linear, 'externalized' sequences, they can also be fruitfully applied to grammatical relations, permitting measurement of the inherent information content of an individual object or operation (Biberauer 2019b, Grünwald 1996, 2007, Newmeyer & Preston 2014). Sequence complexity is identified with richness of content (Mitchell 2009), such that any given signal or sequence is regarded as complex if it is not possible to provide a compressed representation of it. While complexity differences across languages can be measured, for example, as a function of featural specifications on functional elements (Longobardi 2017), here we are interested in the complexity of I-language derivations. Previous efforts have already connected the theory of program size to psychology by implementing a concrete language of thought with Turing-computable Kolmogorov complexity (Romano et al. 2013), which satisfied the requirements of (i) being simple enough so that the complexity of any binary sequence can be measured, and (ii) utilizing cognitively plausible operations like *printing* and *repeating*. In contrast, we aim to relate similar measures to syntactic economy criteria.

The concept of *syntactic complexity* remains underexplored in the literature relative to other measures of linguistic complexity (Shieber 1985, Trudgill 2011). While syntacticians have proposed economy principles, these all effectively boil down to efficient tree-search, staying true to basic intuitions like "less search is preferable to more search"—without formalizing these intuitions or attempting to broach this topic with neighboring fields that might be able to provide highly valuable analytic tools.

Syntactic complexity can be operationalized across a number of dimensions, such as online processing/parsing complexity (Hawkins 2004), tree-search and node counts (Szmrecsányi 2004), number of MERGE applications (Samo 2021), cyclic/derivational complexity (Trotzke & Zwart 2014), internal representational complexity as opposed to derivational size (van Gelderen 2011), entropy reduction (Hale 2016), or stages of second-language development (Walkden & Breitbarth 2019). Syntactic complexity can be framed as grammar-based (derivational), or user-based (parsing) (Newmeyer & Preston 2014); here, will be elaborating on the former type. Crucially, sentence length does not always scale with syntactic complexity (Szmrecsányi 2004), and instead an examination of the underlying operations is required. Although syntactic complexity is often thought of in derivational terms, removed and independent from surface realization, Kolmogorov complexity is relatively theory-neutral and can be applied indiscriminately to mental objects with any number of internal sequences, patterns, irregularities, and surface redundancies (Miestamo et al. 2008). Perhaps most importantly, many postulates within theoretical syntax (traditionally grounded in recursive function theory) cannot be obviously derived from independently understood mathematical, physical or biological constructs, whereas the above complexity metrics currently enjoy a more principled relationship with the natural sciences (Ganguli & Sompolinsky 2012).

Why do we focus here on such a generic, theory-neutral measure? We stress that Kolmogorov complexity (and the related notion of minimal message length) relates directly to frameworks emerging from the FEP (Friston 2019, Hinton & Zemel 1994, Korbak 2021), with the prediction for natural language syntax of reducing the complexity of hierarchical syntactic structures that are read off at conceptual interfaces being sympathetic to a corollary of the FEP that every computation comes with a concrete energetic cost (Jarzynski 1997, Sengupta & Stemmler 2014). As shown above (Eq. 2), variational free energy can be formulated as a trade-off between accuracy and complexity, whereby minimising complexity minimises variational free energy. Considering the topic of universal computation, as in Solomonoff induction (Solomonoff 1964) (which is directly grounded in the minimization of Kolmogorov complexity), many formulations of variational free energy minimization explicitly invoke algorithmic complexity and the type of mathematical formulations underlying universal computation. Relating this more directly to our present concerns, the theme of message length reduction has been fruitfully applied to analyses of grammar acquisition in children. Rasin et al. (2021) show that minimum description length (closely related to Bayesian grammar induction) can provide the child with a criterion for comparing hypotheses about grammatical structures that may match basic intuitions across a number of cases. The *restrictiveness* generated by these complexity measures supplements the more general *simplicity* criterion of theoretical syntax; much as how the 'subset principle' (restrictiveness) supplemented the original evaluation metric (simplicity) (Berwick 1985). Lambert et al. (2021) demonstrated that the computational simplicity of learning mechanisms appears to have a major impact on the types of patterns found in natural language, including for syntactic trees, and so it seems to us well motivated to turn to the issue of the underlying processes that guide the generation of these structures.

Other recent work has successfully used minimum description length in a domain much closer to our own concerns. Focusing on semantic properties of quantifiers (e.g., 'some', 'most') and noting that natural language quantifiers adhere to the properties of *monotonicity*, *quantity* and *conservativity* (Barwise & Cooper 1981), van de Pol et al. (2021) generated a large collection of over 24,000 logically possible quantifiers and measured their complexity and whether they adhered to the three universal properties. They found that quantifiers that satisfied universal semantic properties were less complex and also exhibited a shorter minimal description length compared to quantifiers that did not satisfy the universals, pointing in intriguing directions towards efficiency biases in natural language semantics that appear to restrict the development of lexical meaning. Quantifiers that adhere to semantic universals are *simpler* than logically possible competitors that do not. Do these observations also apply at the level of syntactic structure building?

To briefly formalize our discussion of compression and complexity, given a Turing machine *M*, a program *p* and a string *x*, we can say that the Kolmogorov complexity of *x* relative to *M* is defined by the length of *x*. Formally, this can be expressed as follows (Eq. 4), where $|p|$ denotes the length of *p* and *M* is any given Turing machine:

$$K_M(x) \stackrel{\text{def}}{=} \min\{|p| : M(p) = x\} \cup \{\infty\} \qquad \text{(Eq. 4)}$$

This represents the length of the shortest program that prints the string *x* and then halts. Yet, as implied by Gödel's incompleteness theorem or Turing's halting theorem, we cannot compute the Kolmogorov complexity of an arbitrary string, given that it is impossible to test all possible algorithms smaller than the size of the string to be compressed, and given that we cannot know that the Turing machine will halt (Chaitin 1995). We therefore used an estimate of approximate Kolmogorov complexity (given its fundamental incomputability) based on the Lempel–Ziv compression algorithm, which we applied to the labeling/search algorithm needed to derive each syntactic node in (2) and their subordinated terminal elements, investigating how 'diverse' the patterns are that are present in any given representation of a syntactic tree-search. In the service of replicability and for completeness, we used a current generative grammar labeling/search algorithm that checks tree-structure heads and terminal elements (Chomsky 2013, Ke 2019, Murphy & Shim 2020) (see also f.n. 11). In this respect, Kolmogorov complexity is a more fine-grained measure of complexity than previous measures in theoretical syntax (e.g., node count across a c-commanded probe-goal path).

Searching the structure from top to bottom, identifying each branching node and its elements, we used a Lempel–Ziv implementation (Faul 2021) of the classical Kolmogorov complexity algorithm (Kaspar & Schuster 1987, Lempel & Ziv 1976) to measure the number of unique sub-patterns when scanning the string of compiled nodes[7]. This Lempel–Ziv algorithm computes a Kolmogorov complexity estimate derived from a limited programming language that permits only copy and insertion in strings (Kaspar & Schuster 1987)[8]. The algorithm scans an *n*-digit sequence, $S = s_1 \cdot s_2 \cdot \ldots s_n$, from left to right, and adds a new element to its memory each time it encounters a substring of consecutive digits not previously encountered. Our measure of Kolmogorov complexity takes as input a digital string and outputs a normalised measure of complexity (Urwin et al. 2017).

---

[7] We therefore assume that labeling/search occurs top-to-bottom, and not bottom-up, due to the former yielding a more minimal search path with no 'backtracking', hence more in line with economy considerations.

To connect these ideas with active inference, we note that minimising free energy corresponds to minimising complexity, while maximising the accuracy afforded by internal representations[9] $r \in R$, of hidden states $s \in S$, given outcomes $o \in O$ (Eq. 2). In short, belief updating or making sense of any data implies the minimisation of complexity:

$$D_{KL}[Q(s) \| P(s)] \approx D_{KL}[P(s|o) \| P(s)] \qquad \text{(Eq. 5)}$$

When choosing how to sample data, the expected complexity becomes the intrinsic value or expected information gain (in expected free energy):

$$\mathbb{E}[lnP(s|o,\pi) - lnP(s|\pi)] = I(S,O|\pi) = \mathbb{E}_{P(o|\pi)}[D_{KL}[P(s|o,\pi) \| P(s|\pi)]] \quad \text{(Eq. 6)}$$

This is just the mutual information between (unobservable) hidden states generating (observable) outcomes, under a particular choice or policy.

---

[8] Kaspar and Schuster (1987) discovered that a readily calculable measure of Lempel-Ziv algorithmic complexity can, for simple cellular automata, separate pattern formation from the mere reduction of source entropy, with different types of automata being able to be distinguished. We also note that Lempel-Ziv complexity does not simply measure the number of elements in a sequence, but also factors in pattern irregularities. As such, it is not the case that, by definition, a syntactic process with *n* steps will be trivially more Lempel-Ziv complex than a syntactic process with *n*-1 steps.

[9] That parameterise posterior beliefs $Q(s) \triangleq Q_r(s)$.

Importantly, variational free energy and formulations of artificial general intelligence pertaining to universal computation both share a mathematical legacy. This is rooted in the relationship between the complexity term in variational free energy and algorithmic complexity (Hinton & Zemel 1994, Wallace & Dowe 1999), described in terms of information length and total variation distance. As such, relating syntactic operations to algorithmic compression maximisation feeds directly into assumptions from active inference (Schmidhuber 2010).

Lempel-Ziv complexity is a measure of algorithmic complexity which, under the law of large numbers, plays the same role as the complexity part of log model evidence or marginal likelihood. Interestingly, minimising algorithmic complexity underwrites universal computation, speaking to a deep link between compression, efficiency and optimality in message passing and information processing.

With this background, we now return to the structures in (2b) (licensed) and (2c) (unlicensed). Inputting the labeled nodes (subscript elements) and terminal elements (regular script) across both structures into the Lempel–Ziv compression algorithm (Faul 2021) left-to-right, the licensed representation in (2b) exhibits a normalized Kolmogorov complexity of 1.88, while the unlicensed representation in (2c) exhibits a complexity of 1.99. Crucially, while both (2b) and (2c) exhibit the same node-count complexity and depth (i.e., bracket count), they can be operationally distinguished by their Kolmogorov complexity, in compliance with what the FEP would demand. The increased compression rate for (2b) indicates lower information content, hence lower

Kolmogorov complexity (Juola 2008), and so the representation adheres to the priority to minimise computational load.[10]

### 3.3. Redundancy

Can these types of complexity considerations find a broader architectural basis in theoretical linguistics? Concerns over computational efficiency and representational redundancy motivated the transition beginning in the 1990s from X-Bar models of phrase structure rules involving intermediate hierarchical nodes (e.g., {TP {NP {T' {T {VP {V' {V NP}}}}}}) to Bare Phrase Structure (BPS), which simply involved set-Merge (e.g., {T {NP {V {T {V {V, NP}}}}}}). One of the goals of BPS in abandoning indexes, traces and bar-level representations was to minimize the computational burden postulated both within syntax and during the interface of syntax with language-external interpretive systems. To align these themes with our broader concern of processing dynamics (and, hence, biological plausibility), the psycholinguistic validity and polynomial parseability of BPS minimalist grammars has recently been demonstrated (Torr et al. 2019). BPS attempts to express parsing stages using the fewest number of possible derivational steps, which is not a concern inherent to other forms of parsers. Both X-Bar and BPS models have been represented with rooted, directed acyclic graphs. The main intuitive difference between X-Bar and BPS concerns when node closure is said to occur, and 'how much' node closure occurs (Stabler 1997, Yngve 1960). In addition, while X-Bar theory permits both binary and unary branching, BPS permits only binary branching; every non-terminal node has exactly two daughters. BPS syntax effectively generates simpler trees, yet with a late-stage burden of evaluating the labeled status of a multi-phrase unit and transferring the constructed workspace to the interfaces (Adger 2003).

---

[10] This example of grounding the No Tampering Condition in a theory-neutral measure of algorithmic complexity relates also to another economy condition in linguistics, the Extension Condition. This states that "Merge always applies in the simplest possible form: at the root" (Chomsky 1995: 248), i.e., there is only one site at which to extend a phrase marker. This condition minimizes complexity and search space, and hence aids compressability of structures fed to the interfaces.

### 3.4.  Relativised Minimality

A further observation pertaining to economy in the literature concerns Relativised Minimality (Rizzi 1990, 1991): Given a configuration, [X … Z … Y], "a local relation cannot connect X and Y if Z intervenes, and Z fully matches the specification of X and Y in terms of the relevant features" (Starke 2001). In other words, if X and Y attempt to establish a syntactic relation, but some element, Z, can provide a superset of X's particular features (i.e., X's features plus additional features), this blocks such a relation. In (3a), *which game* provides a superset of the features hosted by *how*, resulting in unacceptability. The equivalent does not obtain in (3b), and so a relationship between both copies of *which game* can be established (strikethroughs denote originally merged positions).

(3)   a.   *[[How$_{[+Q]}$] [C$_{[+Q]}$ [do you wonder [[which game$_{[+Q, +N]}$] [C$_{[+Q]}$ [PRO to play ~~how$_{[+Q]}$~~]]]]]]

     b.   [[Which game$_{[+Q,+N]}$] [C$_{[+Q]}$ [do you wonder [[how$_{+Q}$] [C$_{+Q}$ [PRO to play ~~which game$_{[+Q,+N]}$~~]]]]]]

Relativised Minimality emerges directly from minimal search (Aycock 2020): Consider how when searching for matching features in (3b) the search procedure would 'skip' *how* but find the original copy of *which game*.

We note that the notion of movement 'distance' here is relativised to the specific units across the path, and so even here formulations of Relativised Minimality seem highly specific to language, as opposed to emerging from some kind of domain-general constraint. Hence, in order to reach a more fundamental analysis we may need some means of understanding what 'distance' actually reduces to. Under our approach, algorithmic complexity and compressibility may offer a direction forward, potentially being related in some manner to postulates such as syntactic 'phases'. For example, properties of phases, and phase heads, may fall out of independent features of mental computation, perhaps relating to some threshold of compressibility that serves to distinguish one phase from another when assessed alongside representational category. These speculations aside, we can at least conclude for now that there are avenues of current research that lend themselves quite readily to explorations directed by notions of complexity and compression.

### 3.5. Resource Restriction

The principle of Resource Restriction (or 'Restrict Computational Resources', RCR; Chomsky 2019, Chomsky et al. 2019) states that when the combinatorial operation MERGE maps workspace *n* to workspace *n*+1, the number of computationally accessible elements (syntactic objects) can only increase by one (Huybregts 2019, Komachi et al. 2019). This can account for a peculiar property of natural language recursion that separates it from other forms of recursion (e.g., propositional calculus, proof theory): natural language MERGE involves recursive mapping of workspaces that removes previously manipulated objects (Chomsky 2021c). Hence, Resource Restriction renders natural language derivations strictly Markovian: The present stage is independent of what was generated earlier, unlike standard recursion. Similar observations apply to the idea that when MERGE targets objects in a workspace, non-targeted elements remain conserved and intact once the new workspace has been established (Chomsky et al. 2019), continuing the intuition of 'No Tampering'. MERGE itself exhibits the formal characteristics of a finite-state rewrite rule (Trotzke & Zwart 2014: 145), exhibiting minimal computational complexity, with MERGE being distinct from the ultimate grammatical constructions later derived from its cyclic application.

A topic of recent discussion concerns how we can define the 'size' of a workspace. Fong et al. (2019) suggest that the size of a syntactic workspace should be considered to be the number of accessible terms plus the number of syntactic objects. This proposal to constrain syntactic combinatorics can account for why the applications of certain types of MERGE are ungrammatical (Fong et al. 2019). This theory also requires only element counting—opening up questions about whether other measures may be used to establish legal operations, such as compressibility.

Principles such as Resource Restriction and other economy considerations are essential once we consider that a workspace with two elements with a simple MERGE operation can generate excessive levels of combinatoriality. Within 8 MERGE steps from two elements, around 8 million distinct sets can be formed (Fong & Ginsburg 2018). Older definitions of basic syntactic computations did not "worry about the fact that it's an organic creature carrying out the activities", as Chomsky (2020) notes.

Many aspects of these theories exhibited, to borrow a phrase from Quine (1995: 5), an "excess of notation over subject matter". Even many current models of syntax have ignored questions of cognitive, implementational plausibility (e.g., Chomsky 2013, Citko & Gračanin-Yuksek 2021, Collins 2017, Epstein et al. 2022). Computational tractability (van Rooj and Baggio 2021) is a powerful constraint in this respect (e.g., implementable in polynomial time), and given that minimizing the model complexity term (in formulations of free energy) entails reducing computational cost, this efficiency constraint is also implicitly present in the FEP.

### 3.6.    Interim Conclusion

We have considered how the FEP can in principle provide a novel explanation for the prevalence of efficiency-encoded syntactic structures. To further stress this point, consider Dasgupta and Gershman's (2021) assessment that mental arithmetic, mental imagery, planning, and probabilistic inference all share a common resource: memory that enables efficient computation. Other domains exhibiting computational efficiency include concept learning (Feldman 2003), causal reasoning (Lombrozo 2016) and sensorimotor learning (Genewein & Braun 2014). As Piantadosi (2021) reviews, human learners prefer to induce hypotheses that have a shorter description length in logic (Goodman et al. 2008), with simplicity preferences possibly being "a governing principle of cognitive systems" (Piantadosi 2021: 15; see Chater & Vitányi 2003).

So far, we have only provided one exemplar case of deriving a component of syntactic design from complexity concerns that relate to formulations of variational free energy minimisation. We will now turn to the most commonly explored syntactic processes claimed to arise from economy considerations: syntactic movement and minimal search. Further examples will be used to motivate what we term the principle of Turing–Chomsky Compression, through which stages of syntactic derivations are evaluated based on the algorithmic compressibility of some feature of the computation, such as the movement path of an object, or the procedure of nodal labeling/search—which can be unified based on how they manipulate the syntactic workspace.

### 4. Minimising Free-Energy, Maximising Interpretability

As has long been recognised, the syntactic categories of words are not tagged acoustically, and yet sentential meaning is inferred from syntactic categorization (Adger 2019). Interpreting thematic relations (*who did what to whom*) demands that relations between words are established, however spoken sentences are often ambiguous between distinct syntactic structures. For instance, below we can interpret Jim Carrey as starring in the movie (4a), or sitting next to us (4b).

(4) a.  $[_{TP} [_{NP}$We$] [_{VP}$watched $[_{NP}$a $[_N$movie $[_{PP}$with $[_{NP}$Jim Carrey$]]]]]]$

b.  $[_{TP} [_{NP}$We$] [_{VP}[_{VP}$watched $[_{NP}$a movie$]] [_{PP}$with $[_{NP}$Jim Carrey$]]]]$

Linear distance (i.e., the number of intervening elements between dependents in a sentence) can be contrasted with structural distance (the number of hierarchical nodes intervening), and only the latter is a significant predictor of reading times in an eye-tracking corpus (Baumann 2014). Violations of hierarchical sentence rules results in slower reading times (Kush et al. 2015), and expectations of word category based on hierarchical grammars also predicts reading times (Boston et al. 2011).

The apparent use of hierarchical structure to *limit* interpretation adheres to a core tenet of the FEP, whereby interpretive processes that yield the lowest possible amount of complexity (and thereby computational cost) can mostly (perhaps entirely; Hinzen 2006) be derived directly from what the syntactic component produces. This notion is closely related to the imperatives for structure learning (Tervo et al. 2016)—or Bayesian model reduction—in optimising the structural (syntax) of generative models based purely on complexity (pertaining to model parameters); see Friston, Lin et al. (2017) for an example simulating active inference and insight in rule learning.

While sensorimotor systems naturally impose linear order, linguistic expressions are complex *n*-dimensional objects with hierarchical relations (Gärtner & Sauerland 2007, Grohmann 2007, Kosta et al. 2014, Murphy 2016). The following sections provide concrete demonstrations of these design principles in action.

### 4.1. Structural Distance

Consider the sentence in (5).

(5)        Routinely, poems that rhyme evaporate.

In (5), 'routinely' exclusively modifies 'evaporate'. The matrix predicate 'evaporate' is closer in terms of *structural distance* to 'routinely' than to 'rhyme', since the relative clause embeds 'rhyme' more deeply (minimal search is partly "defined by least embedding"; Chomsky 2004: 109)[11]. Language computes over structural distance, not linear distance (Berwick et al. 2011, 2013, Friederici et al. 2017, Martin et al. 2020).

This can also be shown with simple interrogative structures. Consider the sentence in (6a) and its syntactic representation in (6b), where the verb in the relative clause ('rhyme') is more deeply embedded than 'evaporate'.

(6) a.     Do poems that rhyme evaporate?

b.     [CP[C Do][TP[DP[DP poems][CP[C that][TP rhyme]]][T'[T][V evaporate]]]]

We can compute the complexity of both nodal search and Kolmogorov complexity, contrasting the grammatical association between 'Do' and 'evaporate', and the ungrammatical association between 'Do' and 'rhyme'. When the [+Q] feature on C searches for a goal, it needs to search down three node steps (from CP to V) to get to the grammatical option, but needs to search down four node steps (from CP to embedded TP) to construct the ungrammatical option. Since we are concerned with analyzing a small but representative number of syntactic derivational processes, this analysis differed from the approach to the structures in (2), which did not involve any

―――――――――――――――――

[11] We refer the reader to Ke (2019: 44) and Aycock (2020: 3-6) for a detailed discussion of minimal search, which can be formally defined below, from Aycock (2020), adopting an Iterative Deepening Depth-First Search approach (Korf 1985); where MS = minimal search, SA = search algorithm, SD = search domain (where SA operates), ST = search target:
(1) MS = ⟨SA, SD, ST⟩
(2) SA:
a. Given ST and SD, match against every head member of SD to find ST [initial depth-limit of SD = 1; search depth-first].
b. If ST is found, return the head(s) bearing ST and go to d. Otherwise, go to c.
c. Increase the depth-limit of SD by 1 level; return to a.
d. Terminate Search.

labeling procedure. This time, we enumerated the search steps across nodes, replacing specific nodal categories with symbols interpretable to the Lempel–Ziv compression algorithm (Faul 2021), since this is what the syntactic search algorithm is claimed to monitor. The Lempel-Ziv complexity of the sequence of steps enumerated from the C-V labeling/search algorithm is 1.72. For the embedded C-TP search, it increases to 2.01.

Predictions about grammaticality generated by tree-search depth and Kolmogorov complexity estimates such as Lempel-Ziv may provide an additional advantage to models of syntax that attempt to establish these (and potentially other) forms of sympathies with active inference, relative to rival theories. For instance, while one might invoke purely semantic constraints on polar interrogatives and other forms of question-formation (Bouchard 2021) to derive the kinds of acceptability contrasts we have discussed, we see no way to ground these observations in concerns of computability and complexity, and no way to measure or formalize these notions.

## 4.2.   Ignoring Other People: Question Formation via Economy

As recent literature has explored, whenever there is a conflict between principles of computational efficiency and principles of communicative clarity, the former seems to be prioritized (Asoulin 2016, Murphy 2020a). For instance, consider (7).

(7)      You persuaded Saul to sell his car.

The individual ('Saul') and the object ('car') can be questioned, but questioning the more deeply embedded object forces the speaker to produce a more complex circumlocution ('[ ]' denotes the originally merged position of the *wh*-expression).

(8) a.     *[What] did you persuade who to sell [ ]?

    b.     [Who] did you persuade [ ] to sell what?

The structures in (8) involve the same words and interpretations, yet the more computationally costly process of searching for—and then moving—the more deeply embedded element cannot be licensed, despite the potential benefits of communicative flexibility. Interestingly, one cannot feasibly posit parsing-related factors to derive some independent complexity measure to explain this contrast (e.g., Newmeyer 2007), given the same number of words and same semantic interpretations (i.e., *give me the Agent and Object of the event*). Experimental work has supported the prevalence of these grammaticality intuitions (Clifton et al. 2006). Crucially, this is not to say that when language is used for communication that this process is not also structured via criteria of efficiency; for instance, see Gibson et al. (2019) (but see also Galantucci et al. 2020 for evidence that people often fail to—and, indeed, do not care to—communicate faithfully). Rather, we aim to stress the existence of *conflicting* goals of economy.[12]

Consider also a model of dialogue through which internal generative models used to infer one's own behaviour are deployed to infer the communicative intentions of another, given both parties have similar generative models (Friston & Frith 2015a). As such, the core notion of surprise might be a crucial adjudicator in framing the distinction between I-language and more social conceptions of language production: Reducing surprise/effort internally serves I-language functions of computational efficiency, but it can also serve to encourage mutual predictability between speakers, while, on the other hand, *intentionally inducing surprise* in others can often directly serve certain communicative goals (Giorgi & Dal Farra 2019).

______________________

[12] We also highlight here the generalisation, discussed extensively in Jackendoff and Wittenberg (2014), that simpler syntactic structures typically lead to a greater reliance on pragmatics for successful communication, whereas larger sentences lead to more of an interpretive burden being placed on syntactic principles instead of conversational context.

The syntactic structures for both (8a) and (8b) are represented in (9) (where <DP> represents the movement path). With respect to tree-search depth, (9a) involves searching down 11 nodes, while (9b) involves searching down 9 nodes. To expand our survey of syntactic processes beyond labeling/search paths, we focused here on the postulated path of syntactic object movement across the structure. The movement path was represented with each site being attributed a symbol fed to the compression algorithm, in keeping with a more general approach to annotating movement paths (Adger 2003). Enumerating the movement path from the initially merged root, to intermediate landing sites, to the terminal landing site in Spec-CP, the Lempel-Ziv complexity of movement for (9a) is 2.15. For (9b), path complexity is 1.5.

(9) a.     [CP [DP what] [C' [C did] [TP [DP you] [T' [T *pres*] [vP [<DP>] [v' [v persuade] [CP [C' [C Ø] [TP [<DP>] [T' [T ] [vP [DP who] [PP [P to] [vP [v' [v buy] [<DP>]]]]]]]]]]]]]]]

b.     [CP [DP who] [C' [C did] [TP [DP you] [T' [T *pres*] [vP [<DP>] [v' [v persuade] [CP [C' [C Ø] [TP [<DP>] [T' [T ] [vP [<DP>] [PP [P to] [vP [v' [v buy] [DP what]]]]]]]]]]]]]]]

A further empirical reason to assume that this economy condition is a general property of language comes from the following data of Bulgarian multiple *wh*-fronting (Bošković & Messick 2017; see also Dayal 2017 for discussion of the Superiority Condition). The *wh*-phrase highest prior to movement (the subject in (10) and the indirect object in (11)) needs to be first in the linear order of the sentence, such that the structurally highest *wh*-phrase moves first, and the second *wh*-phrase either right-adjoins to the first *wh*-phrase, or moves to a structurally lower Spec-CP position.

(10)    a.    \*Koj   e      vidjal  kogo?

                *who   is     seen   whom*

          b.    Koj   kogo   e      vidjal?

                "Who saw whom?"

(11)    a.    Kogo  kakvo e      pital   Ivan?

                *whom what   is     asked Ivan*

                "Whom did Ivan ask what?"

          b.    \*Kakvo kogo e      pital   Ivan?

Thus far, this suffices to show that the *wh*-element easiest to search for is selected for movement. However, does syntactic economy simply rule out all but one option? Crucially, Bošković and Messick (2017) show that when multiple options of equal tree-geometric complexity are available, they are *both* licensed as grammatical. Consider constructions with three *wh*-phrases. We can assume that whichever *wh*-element moves to the structurally highest position (Spec-CP) satisfies the featural requirement of interrogative C to have a filled Spec-CP position. After this structurally highest element moves to Spec-CP, we can further assume that the remaining *wh*-elements then move to Spec-CP to satisfy their own featural 'Focus'-based requirements. At this point, whichever order the remaining *wh*-elements move in, the requirements are satisfied through movements of identical length (i.e., both cross the same number of nodes, and hence generate the same sequence of derivational steps, and therefore the same Lempel–Ziv complexity). As such, this predicts that the remaining two *wh*- elements can move in any order after the initial *wh*-movement of the subject. This prediction is borne out: the subject ('koj') is moved first in both constructions below, but then either of the remaining *wh*-elements can move in any order.

(12)     a.     Koj     kogo  kakvo e       pital?

                *who    whom what   is      asked*

                "Who asked whom what?"

         b.     Koj     kakvo kogo  e       pital?

### 4.3.    Subject-Auxiliary Inversion

Another use of minimal tree-search to derive acceptability dynamics is found in the observation that MERGE cannot select the linearly closest auxiliary in English Subject-Auxiliary inversion, but must search for the structurally closest auxiliary.

(13)     Somebody who is in Texas is on the phone.

(14)     a. *[Is] somebody who [ ] in Texas is on the phone?

         b. [Is] somebody who is in Texas [ ] on the phone?

In (14b), the search algorithm is a depth of 3 (moving from the matrix CP to the immediately embedded VP, 'is on the phone', and finding its root auxiliary), while in (14a) the search is a depth of 4 (from the matrix CP to the immediately embedded relative clause, to the VP, and then its root auxiliary). Mapping the search path of (14a) yields a complexity of 2. Mapping the search path of (14b), which is the same as (14a) minus the intermediate relative clause, yields a complexity of 1.58.

### 4.4. Labeling

As a more stringent test, can Lempel-Ziv complexity shed light on cases in which the ungrammatical derivation has *less* structural tree-geometric complexity than the grammatical derivation? Consider the following case from Murphy and Shim (2020: 204). Putting ancillary technical details aside (see Mizuguchi 2019, Murphy & Shim 2020), (15a) is claimed to be ungrammatical because one final necessary operation on the syntactic workspace has not been carried out; namely, merging 'the students' to the structure marked by γ. For expository purposes, we provide a schematic representation to demonstrate the relevant movement path (the path of 'the student' is marked by *t*).

(15)     a. *[$_\gamma$ Seems to be likely [$_\alpha$ the student [to [*t* understand the theory]]]]

         b. [$_\delta$ The student [$_\gamma$ seems to be likely [$_\alpha$ *t* [to [*t* understand the theory]]]]]

The explanations from within syntactic theory as to why (15b) is grammatical concern successful feature valuation and the minimal search of copies via the labeling algorithm. However, this process might also be linked to more efficient compression rates of syntactic labels at the interpretive systems. We can enumerate each labeled node left-to-right marking the phrase boundaries separating each embedded object that pertain to the grammaticality contrast (e.g., V-D-P-V). Computing the Lempel–Ziv complexity of each successive phrase label in these structures, (15a) exhibits a complexity of 1.86, while (15b) exhibits a complexity of 1.66, despite (15b) being a more complex structure from the perspectives of node count and element count. As such, both minimal search of syntactic labels and algorithmic compression rates may be playing separate but related roles in determining how the interpretive systems access objects generated by syntax.

*4.5.    Turing–Chomsky Compression*

The brief number of exemplars we have derived syntactic economy principles from, using a Lempel-Ziv estimate of Kolmogorov complexity, motivate the following principle of language design:

> **Turing–Chomsky Compression**: An operation ($M$) on an accessible object ($O_1$) in a syntactic workspace ($W_p$) minimizes variational free energy if structures from the resulting workspace ($W_q$) are compressed to a lower Kolmogorov complexity than if $M$ had accessed $O_2$ in $W_p$.

This is principally named after specifications over *what* (Chomsky) is compressed and *how* (Turing) such compression can be achieved (Chomsky 2021c, Turing 1950). The interaction between Turing–Chomsky Compression (TCC) and more domain-specific subcategorization requirements emerging from lexico-semantic features, and formal syntactic features, is a promising topic for future research. We have shown across a small but representative number of syntactic processes that derivations minimising algorithmic complexity are licensed over those that result in structures and derivational paths that are less compressible. In keeping with TCC, the examples we presented constitute preliminary evidence that operations on syntactic workspaces are evaluated by criteria that is independent of language-specific representational features—a step towards a "genuine explanation" (Chomsky 2022) for language design.

## 5.  Future Work

> "Language and thought, in anything remotely like the human sense, might indeed turn out to be a brief and rare spark in the universe, one that we may soon extinguish. We are seeking to understand what may be a true marvel of the cosmos."

> Chomsky (2021c: 4)

We have arrived at a number of suggestive explanations for the way language implements the construction of hierarchical syntactic objects: to minimise uncertainty about the causes of sensory data, and to adhere to a least effort natural law (i.e., variational principle of least action), when composing sets of linguistic features for interpretation, planning, prediction and externalization. We have shown that measuring a Kolmogorov complexity estimate of syntactic representations and movement paths can align with acceptability judgments. This was used to motivate a new principle of language design, Turing–Chomsky Compression (TCC). Our use of Lempel–Ziv complexity presents a more explicit measure than previous accounts. For instance, consider Sternefeld's (1997) *Global Economy Condition*, which states that, given two derivations of a syntactic structure (D1, D2), D1 is preferred if D1 fares better than D2 with respect to some metrical measure M (namely, number of derivational steps). This basic 'step counting' measure (as with tree-search depth) seems to be in line with grammaticality predictions of the more general complexity measure provided by Lempel–Ziv complexity. Yet, Lempel–Ziv complexity also benefits from being applicable across a range of other domains in syntax where nodal count does *not* differ between competing structures, and is also related to formulations of variational free energy minimization. Ultimately, this has the advantage of generating quantitative predictions for syntactic computation based on general principles that apply more broadly.

Following neighbouring research in the active inference framework (Da Costa et al. 2021), one could feasibly view our research programme as comparing the information length of belief updating between distinct syntactic derivations and theories. We view our proposals as being, in principle, concordant with the view that neural representations in organic agents evolve by approximating steepest descent in information space towards the point of optimal inference (Da Costa et al. 2021). Future work could explore the utility of minimum description length (van de Pol et al. 2021) and Gell-Mann/Lloyd 'effective complexity'. In contrast to Kolmogorov complexity, which measures the description length of a whole object, effective complexity measures the description length of regularities (structured patterns) within an object (Gell-Mann & Lloyd 1996), which may speak to properties of cyclic, phasal computation in natural language.

More recent work has provided evidence for a mental compression algorithm in humans (termed the Language of Thought chunking algorithm) responsible for parsing very basic, binary sequences, providing evidence that human sequence coding involves a form of internal compression using language-like nested structures (Planton et al. 2021). Dehaene et al. (2022) extend this project to auditory sequences and geometrical shapes. We have effectively extended these ideas further into the domain of natural language syntax, suggesting some common capacity for symbolic recursion across cognitive systems being constrained by compressibility.

While hierarchical complexity in language has clear neurobiological correlates and energetic costs involved that relate to entropy and surprisal measures (Brennan et al. 2016), the neurobiological status of more specific components of syntax that pertain to economy criteria—such as the dynamics of BPS parsing—remains much less clear. As a means to establish more direct, and testable connections between these aspects of syntax and neighbouring domains of the cognitive sciences, we have used a small number of exemplar cases to demonstrate that Lempel–Ziv complexity (i) can align with nodal distance measures; (ii) can provide accurate predictions for legal syntactic objects when nodal distance is identical across legal and illegal syntactic objects; and (iii) can implement a principle of free energy minimization. In addition, Lempel–Ziv complexity has been used to successfully measure the complexity of cortical activity across a number of domains, including sleep and wakefulness (Abásolo et al. 2015), the quantification of spikes and bursts of synchronization (Blanc et al. 2008), the aging brain (Shumbayawonda et al. 2018), cortical complexity under general anesthesia (Puglia et al. 2021), and the reduction of resting-state neural complexity in schizophrenia relative to neurotypical controls (Fernández et al. 2011). For linguists, we believe that the complexity metrics we have discussed can constitute supporting hypotheses that connect theories of grammar to observable psychophysical measurements and neural signals.

Some linguists might object to our complexity measure in the following way: Why should syntax be organized so as to produce structures that minimise Kolmogorov complexity, and why should the semantic component of language aim to read off structures that are of corresponding complexity? We note in response that the core 'phase'/non-phase pattern of syntactic derivations (e.g., {C {T $v$ {V D/$n$ {N}}}}; Richards 2011, Uriagereka 2012) optimizes compression rate (effectively, 010101), and since phase construction constitutes the major determining period when syntactic workspaces are accessed by the conceptual systems, we see our proposal as aligning closely with existing—if only implicit—assumptions.

Turning to other alternatives, innovative accounts that have invoked category theory to model natural language syntax (Coopmans et al. 2021) have—thus far—only modeled binary set-formation and lexical selection criteria without using the tools of category theory to explain the design of these syntactic structures, economy criteria, or other more complex syntactic features such as displacement. While category theory is functionally useful for contrasting language with other cognitive faculties such as action (as shown in Coopmans et al. 2021), the separate goal—and our present goal—of grounding apparent economy constraints in syntax within a biologically plausible model of language and computation in neural systems, and how these complex systems implement these constraints, may require other formal tools.

Plainly, there are many issues with the framework we have outlined here that need to be further unpacked and clarified. Our proposals concerning compression of structures accessed from the syntactic workspace via TCC have been discussed in the context of a cursory overview of the mathematical lineage shared between formulations pertaining to the FEP and theories of universal computation. This suited our current expository purposes, with our proposals being buttressed by conceptual overviews of the FEP and syntactic economy, but future work should more carefully align models of active inference with TCC. Although we made our assumptions about syntactic complexity based on whether or not our measure can be formally grounded within the FEP, we note that we have effectively equated complexity with compressibility. As such, we acknowledge that there may be a number of other fruitful directions to measure complexity in ways that are sympathetic to active inference (e.g., the "complexity equals change" framework; Aksentijevic & Gibson 2012).

We also believe that entropy measures (Hale 2016) are closely aligned with our proposals, reflecting internal aspects of parser states, with entropy measuring their uncertainty and relating to expected payoffs, and Kolmogorov complexity measuring their compressibility (indeed, Shannon's entropy is an upper bound on Kolmogorov complexity). For instance, Yun et al. (2015) combine minimalist syntax with an entropy reduction complexity metric to derive word-by-word predictions across Chinese, Japanese and Korean. Looking at different types of relative clauses (subject relative and object relative) that involve different syntactic movement paths, their entropy reduction pattern directly matches independent psycholinguistic measures of reading difficulty. Conceptually speaking, entropy reduction measures and Kolmogorov complexity measures speak to the costs of endogenous syntactic computations, being more directly informative about internal states than more externalist measures like surprisal (Murphy 2021).

A challenge for future research is to appropriately frame the interfacing of syntax with other systems in terms that accord with the minimisation of surprise and variational free energy. Since the FEP has attendant process theories (e.g., active inference), one of the latent payoffs of our suggestions here is the development of generative models of active inference that fully ground specific factors in syntactic theory and, through simulation work, may align with recent advances in the electrophysiology, neural dynamics and neural harmonics of syntax (Keitel et al. 2017, Tavano et al. 2021). For example, one crucial factor in any syntax model seems to be phrasal category information, since it appears to drive cortical tracking of hierarchical structures (Benítez-Burraco & Murphy 2019, Burroughs et al. 2021, Murphy 2020b, Murphy, Hoshi et al. 2022, Wilkinson & Murphy 2016), in line with work highlighting the unique computational contribution of phrase structure labeling (Adger 2013, Hornstein 2009, Murphy 2015b, Zhao 2021).

## 6. Conclusions

We have reviewed how the FEP, that underwrites active inference, is an expression of the principle of least action, which is additionally a principle implemented in models of theoretical syntax. The FEP provides suggestions for *why* the brain computes the way it does, while theoretical linguistics provides a means of establishing *what* is computed. Our discussion of this topic culminated in a novel design principle for natural language, Turing–Chomsky Compression (TCC). An intuition from 70 years ago—that the extent to which a model of grammar is simple seems to determine its explanatory power—echoes in the modern framing of syntactic computation as a system of economy: "[T]he motives behind the demand for economy are in many ways the same as those behind the demand that there be a system at all" (Chomsky 1951; see also Goodman 1951). For Rissanen (1989), "given some data, the simplest hypothesis explaining it is preferable". Generative linguistics has long appealed to economy considerations (e.g., the evaluation metric in Chomsky & Halle 1968). This appeal to simplicity may even emerge from an evolved bias shared across our species to favour simple solutions (see the 'cognitive simplicity hypothesis'; Terzian & Corbalán 2021). The FEP has produced formal, simulation-supported models of complex cognitive mechanisms such as action, perception, learning, attention and communication, while theories of syntax embracing computational efficiency have led to empirically successful outcomes, explaining grammaticality intuitions (Adger 2003, Martin et al. 2020, Sprouse 2011, Sprouse & Almeida 2017), certain poverty of stimulus issues (Berwick et al. 2011, Crain et al. 2017, Culbertson et al. 2012, Wexler 2003, Yang et al. 2017) and the pervasive organizational role that hierarchy has in language (Friederici et al. 2017, Grimaldi 2012) and the seemingly unique proclivity humans have to parse sequences into tree-structures.

Our approach also mirrors previous applications of the FEP to cognition. Exploring the issue of subgoaling, Maisto et al. (2015) conclude that favoured subgoals are ones that permit planning solutions and controlling behaviour using fewer information resources (yielding parsimony in inference and control), and conclude that of those strings that represent procedures returning the same output, strings with lower descriptive complexity are more probable, and should more likely be selected by probabilistic inference. They argue that this framing is compatible with the FEP, and we have further argued here that these assumptions of computational efficiency in subgoals can be found in syntactic combinatorics (e.g., Momma & Phillips 2018 discuss linguistic parsing and production in terms of discrete subgoals). We find seeds for these ideas in the foundational principles of universal computation, where, as we have noted, free energy is often discussed in terms of minimum description or message lengths (MacKay 2003, Schmidhuber 2010). Relatedly, core findings from dependency length minimization (DLM) research suggest that, during online parsing, comprehenders seek to minimize the total length of dependencies in a given structure since this reduces working memory load (Gibson et al. 2019). These overt, 'visible' and linearized relations may therefore be constrained by similar efficiency criteria of the kind we have discussed here, which seem to shape the operations of syntactic structure building.

A core objective of current theories of syntax is to explain *why* language design is the way it is (Adger 2019, Narita 2014), and we have suggested that the FEP can contribute to this goal. The more efficiently a language user can internally construct meaningful, hierarchically organized syntactic structures, the more readily they can use these structures to contribute to the planning and organization of action, reflection and consolidation of experience, exogenous and endogenous monitoring, imagination of possible states, adaptive environmental sampling, and the consideration of personal responsibilities. We used a brief number of examples to demonstrate proof of concept for how compression algorithms, such as a Kolmogorov complexity estimate, can provide principled insight into efficiency concerns alongside more traditional economy criteria such as node count and tree-search depth. Given that one can characterize major components of mental life as being derived from information compression (e.g., casting aside the vast majority of perceptual experience that does not inform generative models), our conclusion that the most basic component of natural language executes the same process as a primary constraint on its activities and its interfacing with other cognitive systems should be seen as closely sympathetic to algorithmic information theories of cognition and consciousness (Ruffini 2017). Other potential topics to explore include Case systems of natural language, which do not seem to be motivated purely by semantic demands (Chomsky 2020, 2021c) and which seem instead to help the facilitation of perception and parsing by establishing clear relations between discourse elements. How this and other components of natural language relate to the priorities of active inference is a promising direction to pursue. Indeed, it may transpire that developing a general understanding of computational efficiency for organic, mental systems may only be possible by adopting an interdisciplinary, systematic approach to natural phenomena (Cipriani 2021), of the kind afforded by the active inference framework, moving beyond the types of descriptions initially tailored specifically to linguistic categories.

Using examples involving minimal search and syntactic movement, we have shown that natural language syntax and its interfacing with conceptual systems is uniformly constrained by rates of algorithmic compression, motivating what we term the TCC principle, which may be a foundational organizing constraint on language design. While linguists can speak of the 'Superiority Condition', the 'labeling algorithm' and 'minimal search', we have claimed that some (perhaps all) of these postulates can be derived from the goal of minimizing variational free energy via TCC. From the perspective of active inference, individuals need to minimise the effort involved in meaning-making, and TCC contributes to this goal. Active inference assumes an imperative to find the most accurate explanation for sensory observations that is minimally complex—recruited in Barlow's (2001) exploration of minimum redundancy—and which seems in accord with how the language system builds structure, and structural relations (Hauser & Watumull 2017).

We note in this context recent work by van Gelderen (2018, 2021). Using a series of historical reviews of extensive language changes, van Gelderen (2021) shows that syntactic economy principles can account for language variation across a broad variety of phenomenon, including *that*-trace constructions, CP-deletion, and the presence of expletives. In addition, van Gelderen (2018) argues that regular patterns of language change can be seen as resolutions to labeling failures (i.e., when the labeling/search algorithm 'fails' to label a given structure). This work reveals how considerations of economy (that we have argued arise directly from active inference) are not simply formal stipulations useful in accounting for grammaticality—they can also yield principled insights into large-scale dynamics in historical language change.

Moving beyond, we note that our measures of Kolmogorov complexity in the operations of natural language syntax are exploiting a highly general, theory-neutral measure of complexity, and that these and other related measures serve to index some underlying, independent process of organic computation, which remains elusive in its formal character and neural basis. While some domain-specificity will continue to be required in the language sciences, if only due to the closed class of lexical objects that syntax accesses, we hope to have shown that theory-neutral measures of complexity that have previously been related to formulations of the FEP can provide new inroads for the study of language.

More broadly, what the FEP can offer theoretical linguistics is proof of principle: a foundational grounding and means of additional motivation for investigating language in terms of efficient computation. The FEP is fundamentally a normative model (Allen 2018) which can aid the generation of implementational/process models (such as predictive coding or active inference) and can place constraints on feasibility. Further simulation and modeling work is required to push these ideas further for natural language and its various sub-systems, and we envisage that this type of future work will provide fruitful insights into natural language syntax. If this project can be expanded in this manner, we could make significant gains in ways that traditional experimental psycholinguistics and neuroimaging of language processing have often been unable to—we may discover a means to reconcile universal properties of human language syntax with endogenous properties of the brain.

***Open Practices Statement***: No empirical data is associated with this article.